\documentclass[letterpaper, 10 pt, conference]{ieeeconf}

\IEEEoverridecommandlockouts
\usepackage{amsmath}
\usepackage{amssymb}
\usepackage{booktabs}
\usepackage{graphicx}
\usepackage{tabularx}
\usepackage{array}
\usepackage{makecell}
\usepackage{multirow}
\usepackage{xcolor}
\usepackage{mathtools}
\let\labelindent\relax
\usepackage{enumitem}
\usepackage{balance}
\usepackage{adjustbox}
\usepackage{stfloats}
\usepackage{subcaption}

\usepackage[backend=biber,style=ieee,natbib=true,url=false,eprint=false,doi=false]{biblatex}

\usepackage{hyperref}
\usepackage[capitalize,noabbrev]{cleveref}

\definecolor{customRed}{RGB}{153,27,30}
\definecolor{customGreen}{RGB}{70,170,80}
\hypersetup{
	colorlinks=true,
	linkcolor=customGreen,
	filecolor=magenta,
	urlcolor=customRed,
	citecolor=customRed,
	pdftitle={Reward-DAgger: Robot-Gated Interactive Imitation Learning with General-Purpose Progress-Based Reward Models},
	pdfauthor={Ryan Li, Yigit Korkmaz, Erdem Biyik},
}

\newcolumntype{Y}{>{\centering\arraybackslash}X}

\definecolor{thrifty}{HTML}{6FAF91}   %
\definecolor{diff}{HTML}{958BC0}      %
\definecolor{ucf}{HTML}{9AA56F}       %
\definecolor{safe}{HTML}{80BBD2}      %
\definecolor{logpzo}{HTML}{BE8D96}    %
\definecolor{reward}{HTML}{D97706}    %

\newcommand{\thriftycolor}[1]{\textcolor{thrifty}{#1}}
\newcommand{\diffcolor}[1]{\textcolor{diff}{#1}}
\newcommand{\ucfcolor}[1]{\textcolor{ucf}{#1}}
\newcommand{\safecolor}[1]{\textcolor{safe}{#1}}
\newcommand{\logpzocolor}[1]{\textcolor{logpzo}{#1}}
\newcommand{\rewardcolor}[1]{\textcolor{reward}{#1}}

\newcommand{\method}{Reward-DAgger}

\newcommand\blfootnote[1]{%
  \begingroup
  \renewcommand\thefootnote{}\footnote{#1}%
  \addtocounter{footnote}{-1}%
  \endgroup
}

\title{\LARGE \bf
\method: Robot-Gated Interactive Imitation Learning\\with General-Purpose Progress-Based Reward Models
}

\author{Ryan Li$^{*1}$, Yigit Korkmaz$^{*2}$, Erdem Bıyık$^{2}$ \\
    $^{1}$Department of Mathematics, University of Southern California \\
    $^{2}$Thomas Lord Department of Computer Science, University of Southern California
}

\begin{document}

\maketitle

\thispagestyle{empty}
\pagestyle{empty}

\begin{abstract}

Recent advances in robot learning have enabled generalist control policies capable of completing a wide range of tasks. However, their performance degrades when deployed in unseen environments, making it critical to detect failures and teach recovery behaviors. 
Existing runtime monitoring methods often require task- and policy-specific training or hyperparameter tuning, limiting cross-task deployment and introducing additional overhead during iterative policy updates.
We present \method{}, a robot-gated interactive imitation learning framework that uses dense progress signals from a general-purpose reward model to determine when human intervention is needed. Our approach is agnostic to the underlying policy architecture, requires no access to policy internals, and can be applied across tasks without retuning the gating mechanism. Our results show that \method{} achieves a better failure-detection accuracy--latency tradeoff than existing runtime monitoring baselines. Across eight simulated and real-world tasks, \method{} consistently improves the downstream policy's autonomous success rate throughout interactive learning and achieves strong return on human effort, outperforming the baselines in most settings. Importantly, the same gating configuration is used across tasks without task-specific hyperparameter tuning, demonstrating transfer across tasks, environments, and policy architectures. Code and videos are available at \href{https://liralab.usc.edu/reward-dagger}{https://liralab.usc.edu/reward-dagger}.

\blfootnote{$^\star$Equal Contribution}
\end{abstract}

\section{Introduction}

Modern learning-based robot policies, ranging from task-specific models \citep{chi2023diffusion, zhao2023learning} to generalist vision-language-action (VLA) models \citep{kim2024openvla, black2024pi0visionlanguageactionflowmodel, octo2024octo}, are increasingly capable of completing challenging manipulation tasks. Yet, these policies remain brittle during deployment: a missed grasp, a dropped object, or a misaligned insertion might cause an otherwise successful rollout to fail. Interactive imitation learning methods, such as DAgger \citep{ross2011reduction} and its human-gated variants \citep{liu2023robot, korkmaz2025mile, kelly2019hgdagger, mandlekar2020humanintheloopimitationlearningusing,peng2025data}, address this problem by collecting expert interventions from states where the policy struggles and using these corrections to iteratively improve the policy. However, continuously monitoring robot execution and deciding when to intervene places a substantial burden on the human expert, making such approaches difficult to scale to robot fleets.

Robot-gated interactive imitation learning instead seeks to automate this decision by requesting human assistance when the policy is likely to fail, thereby making more efficient use of costly human supervision. Existing robot-gating methods, however, commonly rely on either learned failure-prediction functions trained from explicit failure data \citep{hoque2021thriftydagger} or policy-dependent uncertainty scores that require task- and policy-specific calibration \citep{gu2025safe, lee2025diff, he2026uncertainty, xu2025can, myers2023active}. The former requires collecting failures that are undesirable or infeasible during safety-critical deployment, while the latter couples the gating mechanism to a particular policy checkpoint. As the policy changes throughout interactive learning, the gate needs to be retrained or recalibrated, limiting its scalability.

Ideally, a scalable gating mechanism for interactive imitation learning should satisfy four properties: \textbf{(1)} detect failures early while avoiding unnecessary interruptions of successful behavior; \textbf{(2)} remain effective across iterations of interactive learning without repeated retraining or recalibration; \textbf{(3)} operate with different, potentially black-box, policy architectures; and \textbf{(4)} transfer across tasks without requiring a separate training or calibration procedure for each task.

Our key insight is that general-purpose progress-based reward models, which estimate task progress directly from visual observations and natural-language instructions \citep{liang2026robometer, tan2026robodopamine, lee2026roboreward}, provide a natural signal for such a gate. Unlike policy-dependent uncertainty measures, task progress describes what the robot has accomplished rather than how the policy generates its actions. Consequently, progress can be evaluated independently of the underlying policy architecture or checkpoint. This decoupling allows the gating mechanism to be fixed as the policy is updated and to be applied to different policies without access to their internal representations. Moreover, by leveraging reward models pretrained across diverse tasks, progress provides a common signal that can support gating across tasks without task-specific gate training.
Importantly, even under task distribution shift, the reward model only determines when to request assistance; the policy is still trained from human corrections, which provide valid task-specific supervision.

Based on this insight, we introduce \textbf{\method{}}, a robot-gated interactive imitation learning framework that uses a simple but effective two-timescale rule over progress estimates from a general-purpose progress-based reward model to trigger human takeover. A short-window gate detects rapid decreases in progress, capturing immediate failures such as dropping an object, 
while a long-window gate detects progress plateaus corresponding to stalled behavior, such as repeatedly failed grasps. When either gate fires, control is transferred to a human expert, and the resulting corrections are used in the next policy update. Importantly, both the reward model and the gating configuration remain fixed throughout interactive learning, requiring no per-round retraining or recalibration.

We evaluate \method{} along three dimensions: failure-detection performance, cross-task transfer of the gating parameters, and downstream policy improvement. On four simulation tasks, \method{} achieves a strong accuracy--latency trade-off compared to existing robot-gating and runtime-monitoring baselines. We further find that a single gate configuration selected jointly on multiple source tasks transfers to held-out target tasks without reducing detection accuracy and with only a minor increase in detection latency. In simulation, \method{} consistently improves autonomous success rate across three rounds of interactive learning while achieving the highest return on human effort. Finally, we deploy the same simulation-selected gate configuration on four real-world manipulation tasks with large pretrained VLA policies. Without any task-specific hyperparameter tuning, \method{} achieves competitive performance across all tasks and outperforms the baselines in most settings in terms of autonomous success rate and return on human effort, demonstrating that the gating mechanism transfers across environments and policy architectures.

In summary, our main contributions are:
\begin{itemize}[leftmargin=*]
\item We propose a two-timescale, progress-based gating mechanism that detects both abrupt failures and stalled behavior during robot execution. The gate is independent of the underlying policy, remains fixed across policy updates, and supports cross-task configuration transfer.
\item We introduce \method{}, an interactive imitation learning framework that uses this gating mechanism to trigger human takeover and iteratively finetune the policy from the resulting expert corrections.
\item We extensively evaluate \method{} across four simulated and four real-world manipulation tasks with both task-specific diffusion policies and large pretrained VLAs, studying failure detection, cross-task transfer, downstream policy improvement, and human-effort efficiency.
\end{itemize}

\begin{figure*}[t]
    \centering
    \includegraphics[width=0.9\textwidth]{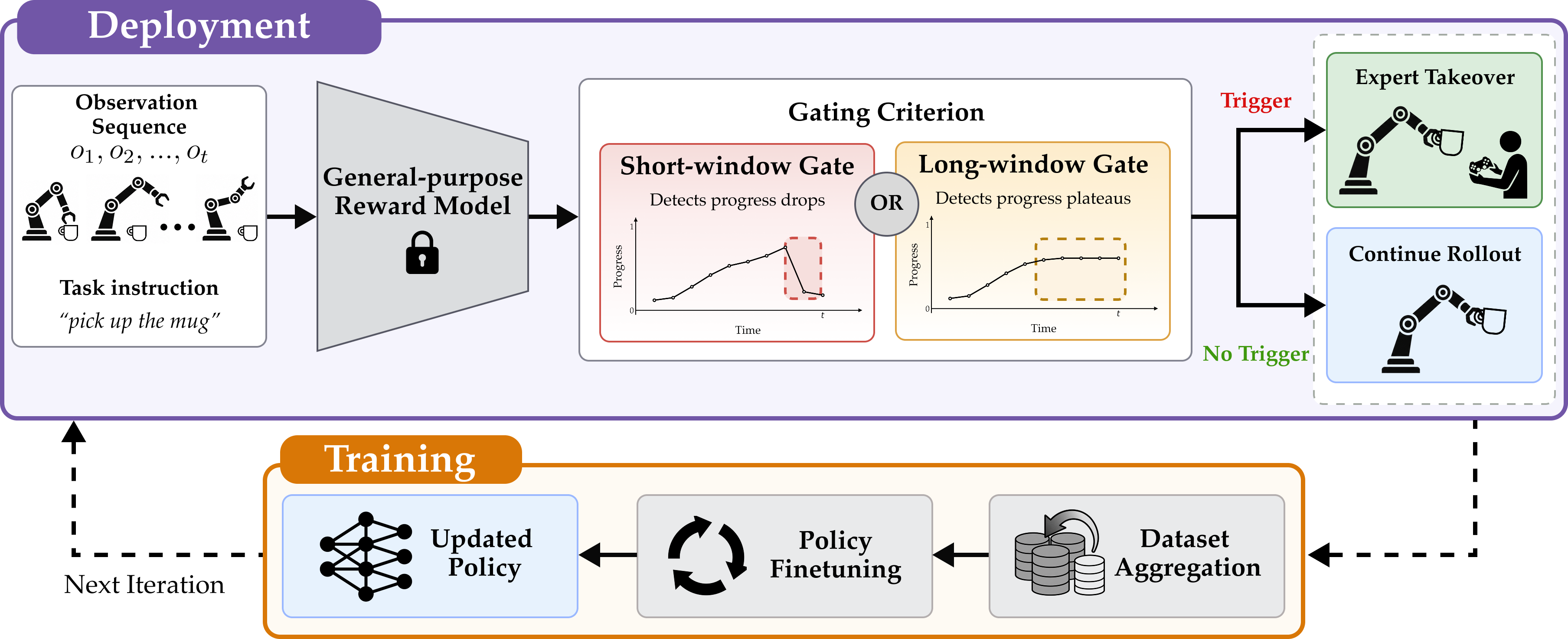}
    \caption{\textbf{\method{}} is a robot-gated interactive imitation learning framework that uses a two-timescale rule over progress estimates from a general-purpose reward model to trigger human takeover. Collected corrections are appended to an initial dataset and the policy is finetuned in an iterative fashion.}
    \label{fig:framework}
    \vspace{-5px}
\end{figure*}

\section{Related Work}

\noindent\textbf{Interactive Imitation Learning.}
Interactive imitation learning mitigates distribution shift in behavioral cloning by iteratively collecting expert actions on states visited by the learned policy \citep{baraka2026humaninteractive, ross2011reduction, kelly2019hgdagger, hoque2021thriftydagger, liu2023robot, korkmaz2025mile}. In \emph{human-gated} methods, a supervisor monitors execution and decides when to intervene \citep{kelly2019hgdagger, liu2023robot, korkmaz2025mile, hu2025rac,peng2025data}. This monitoring requirement becomes burdensome when supervision scales to many robots \citep{liu2024multi, hoque2022fleetdagger, yu2026armada}. \emph{Robot-gated} methods instead automate this decision using signals such as action disagreement, estimated risk, or policy uncertainty \citep{hoque2021thriftydagger, lee2025diff, he2026uncertainty, liu2024modelbased, myers2023active,cai2025robot}. However, these gates commonly depend on task-specific learned criteria or policy-dependent signals that require additional training or calibration across tasks or policy updates. In contrast, \method{} uses an external task-progress signal that can be applied zero-shot across tasks and policy architectures.

\smallskip
\noindent\textbf{Runtime Monitoring in Robot Manipulation.} 
Runtime monitoring provides a closely related approach for identifying policy failures during deployment. These methods typically reduce observations, predicted actions, or internal policy representations to a scalar score, and then calibrate a conformal threshold for failure detection. Representative methods derive these scores from temporal inconsistencies among consecutive actions \citep{agia2024unpacking}, deviations from the nominal observation distribution \citep{xu2025can, romer2025fiper}, action chunk entropy \citep{romer2025fiper}, or learned failure predictors using internal VLA features \citep{gu2025safe}. However, these methods are primarily studied as deployment-time mechanisms without updating the monitored policy. Reward-DAgger closes this loop between monitoring and learning by using gate firings to collect expert corrections for subsequent policy updates without inducing expensive retraining or recalibration overheads.

\smallskip
\noindent\textbf{General-Purpose Reward Models.}
Reward functions provide scalar feedback about the desirability of states or actions and are commonly used to guide policy optimization. Recently, significant effort has focused on developing general-purpose reward models that provide dense reward signals across diverse tasks \citep{chen2026topreward, liang2026robometer, lee2026roboreward, huang2026rynnvalue, zhang2025rewind, sontakke2023roboclip}. A common approach is to estimate task progress from visual observations and language instructions and use this progress signal as a proxy for per-timestep reward \citep{huang2026rynnvalue, liang2026robometer, tan2026robodopamine, zhang2025rewind}, with recent works demonstrating its use for reinforcement learning in real-world robotic manipulation. Despite these advances, directly optimizing policies with imperfect learned reward functions remains challenging, as they may exploit inaccuracies or spurious correlations in the learned reward, leading to reward hacking without completing the intended task \citep{tian2026position, schroeder2026soler1, zhang2026progressrewardmodelingrobotic, tien2023causal}. Rather than directly optimizing the progress signal, our key insight is to use it as a runtime monitoring signal for interactive imitation learning. Since progress is evaluated independently of the underlying policy and represented on a shared scale across tasks, it provides a natural signal for determining when expert intervention is needed without additional task- or policy-specific training or calibration.

\section{Method}

\method{} leverages dense progress signals obtained from general-purpose reward models to determine when a robot policy should transfer control to the expert. In this section, we first introduce how we construct a gating criterion from general-purpose reward models, and then discuss how we incorporate the criterion into an interactive imitation learning pipeline for fine-tuning the base policy.

\subsection{Gating Criterion}

A general-purpose reward model is typically constructed by prompting or fine-tuning a VLM to produce dense progress signals. Formally, let $\ell$ denote a natural-language task instruction and let $o_{\leq t}=(o_1,\ldots,o_t)$ denote the visual observations available up to
time $t$. A general-purpose progress-based reward model $f_\psi$ produces a scalar estimate of current task progress,
\begin{equation}
    \phi_t = f_\psi(o_{\leq t},\ell),
    \label{eq:progress-estimate}
\end{equation}
where larger values indicate greater predicted progress.

Inspired by \citep{liang2026robometer}, we target two common temporal signatures of policy failure in terms of the predicted progress signal. First, the policy may suddenly undo previously achieved progress. For example, a grasped object may slip from the gripper, or a partially inserted object may be pulled back out. These failures produce rapid decreases in predicted progress. Second, the policy may become stuck without advancing the task. For example, it may repeatedly miss a grasp or oscillate near the target pose without completing the placement. Such failures appear as plateaus in the predicted progress.

Following this observation, we design two complementary gates that constitute the overall \emph{reward gate}. At time step $t$, the reward gate has access to the progress history $\phi_{1:t}=(\phi_1,\ldots,\phi_t)$. The \emph{short-window gate} examines the most recent $L_s$ progress estimates, $\phi_{t-L_s+1:t}$, to detect sharp decreases in progress. The \emph{long-window gate} examines the most recent $L_l$ progress estimates, $\phi_{t-L_l+1:t}$, to detect progress plateaus, where $L_l>L_s$.

To quantify drops and plateaus, we use Spearman rank correlation: given a time-indexed window $\mathbf{v}=(v_1,\ldots,v_n)$ of scalar values, we rank its values and compute the Pearson correlation between these ranks and the time indices $(1,\ldots,n)$. We denote the resulting coefficient by $\rho(\mathbf{v})$. A coefficient near $+1$ indicates monotonically increasing progress, a coefficient near $-1$ indicates monotonically decreasing progress, and a coefficient near $0$ indicates no clear monotonic trend. We compute the Spearman correlation within the short and long windows as
\begin{equation}
\rho_t^s=\rho(\phi_{t-L_s+1:t}),\quad 
\rho_t^l=\rho(\phi_{t-L_l+1:t}).
\label{eq:window-correlations}
\end{equation}

One problem with Spearman correlation is that it captures the direction of a progress trend but not the magnitude of the change. This is particularly important for the short-window gate, since a small fluctuation may produce a strongly negative correlation. Therefore, we additionally define the maximum drop magnitude over the window as
\begin{equation}
\Delta_t = \max_{j\in\{t-L_s+1:t\}} \phi_j-\phi_t.
\label{eq:drop-magnitude}
\end{equation}

Using the window correlations and maximum drop magnitude, we define the complete gating criterion as
\begin{align*}
g_t^s=\mathbb{I}\!\left[
(\rho_t^s<\tau_s)\land(\Delta_t\geq\delta_{\min})
\right],
\quad
g_t^l=\mathbb{I}\!\left[\rho_t^l<\tau_l\right],\nonumber
\end{align*}\\[-35px]
\begin{align}
g_t=g_t^s\lor g_t^l\,,
\label{eq:reward-dagger-gate}
\end{align}
where $\tau_s$ and $\tau_l$ are the respective correlation thresholds, while $\delta_{\min}$ sets the maximum drop magnitude threshold for the short-window gate. The reward gate fires when the short-window gate or long-window gate fires. A gate configuration is specified by $(L_s,\tau_s,\delta_{\min},L_l,\tau_l)$. We set $\tau_s<\tau_l$ so that the short-window gate detects progress drops, while the long-window gate detects progress plateaus. Each gate is evaluated only after its corresponding window is full.

\subsection{\method{}}
We now summarize the \method{} procedure. We start with a robot policy $\pi_0$ and an offline dataset of expert demonstrations. During a rollout, the robot executes actions in the environment until the reward gate fires, after which the expert takes over and provides corrections. At each iteration $i$, we collect expert correction transitions until a fixed budget is reached. We then aggregate these expert interventions with the existing dataset and fine-tune $\pi_i$ to obtain the next student policy $\pi_{i+1}$. This process is repeated for a fixed number of iterations without changing the reward gate's configuration. An overview of the framework is shown in \Cref{fig:framework}.

\section{Experiments}

Our experiments evaluate the effectiveness of \method{} as a robot-gated interactive imitation learning framework across different tasks and policy architectures. Specifically, we seek to answer the following research questions:

\begin{enumerate}[label=\textbf{Q\arabic*}., nosep, leftmargin=0pt, labelindent=0pt, itemindent=*]
\item \label{q1} Does the gating mechanism of \method{} accurately detect failures with low detection latency?
\item \label{q2} Does the gating configuration of \method{} transfer across tasks without additional calibration?
\item \label{q3} Does \method{} efficiently improve downstream policy performance while minimizing human burden?
\end{enumerate}

\smallskip
\noindent\textbf{Baselines.}
We compare \rewardcolor{\textbf{\method{}}} against robot-gated DAgger methods as well as recent runtime monitoring approaches adapted to interactive imitation learning:

\begin{itemize}[leftmargin=*]
\item\textbf{\thriftycolor{ThriftyDAgger}} \citep{hoque2021thriftydagger} trains a policy ensemble to estimate state novelty, and a $Q$-function to estimate risk. The human takeover is triggered when a state is deemed novel or risky.

\item\textbf{\diffcolor{Diff-DAgger}} \citep{lee2025diff} uses the diffusion policy’s denoising loss for the proposed action as an uncertainty score, which is then thresholded to trigger human takeover.

\item\textbf{\ucfcolor{UCF}} \citep{he2026uncertainty} samples candidate end-effector poses within a radius of the current pose, fits a Gaussian mixture model to the denoising vectors, and combines inter-mode divergence with intra-mode variance to estimate uncertainty. 

\item\textbf{\safecolor{SAFE-LSTM}} \citep{gu2025safe} trains an LSTM on the final-layer features of a VLA policy to produce a scalar failure score, which is thresholded for detecting failures.

\item\textbf{\logpzocolor{logpZO}} \citep{xu2025can} trains a flow matching model to map in-distribution observations to the standard normal distribution, and uses the distance between the current observation’s latent representation and the origin as the score.
\end{itemize}

\noindent\textbf{Reward Model.}
We instantiate the reward model in \method{} using Robometer-4B~\citep{liang2026robometer}, a general-purpose reward model that produces dense, frame-level estimates of task progress on a normalized $[0,1]$ scale. We select Robometer because it outperforms prior general-purpose reward-model baselines across cross-domain robotic reward evaluations.

\noindent\textbf{Simulation Setup.} For our simulation experiments, we use LIBERO-10~\citep{liu2023libero}, a suite of 10 long-horizon tasks with diverse settings. Simulation enables us to evaluate different methods across multiple tasks at scale, collecting enough rollouts to reliably characterize their accuracy--latency trade-offs. In these experiments, we use a frozen $\pi_{0.5}$~\citep{intelligence2025pi05visionlanguageactionmodelopenworld} LIBERO checkpoint as the expert policy. The student is a diffusion policy~\citep{chi2023diffusion} trained on LIBERO demonstrations.

We first identify Tasks 0, 1, 4, 5, 6, and 8, where the initial student policies show intermediate performance, yielding sufficient numbers of both successful and failed trajectories for meaningful evaluation. For failure-detection and downstream interactive learning experiments, we use Tasks 0, 1, 4, and 6, as these are the tasks our expert policy consistently recovers from frequent failure states. For experiments that only require student-policy rollouts, including hyperparameter selection and cross-task transfer, we use all tasks. 

\noindent\textbf{Real Robot Setup.} Our real-robot experiments use the DROID~\citep{khazatsky2024droid} setup, with a Franka FR3 robot arm equipped with a Robotiq 2F-85 gripper. We evaluate \method{} on four tasks: \textit{stack} the red block on top of the blue block, \textit{pick} up the red marker and put it into the blue cup, \textit{put} the slice of bread from the toaster to the bowl, and \textit{remove} the towel and screwdriver from the basket and put them on the table. In these experiments, the student is a pretrained $\pi_{0.5}$ policy, and the expert is a human teleoperator. An overview of the simulated and real-world tasks is provided in \Cref{fig:task-overview}.

\subsection*{Q1: Failure Detection}
We first evaluate the effectiveness of \method{}'s gating mechanism on individual tasks based on two metrics: 
\begin{itemize}[leftmargin=*]
    \item \textbf{Accuracy}. The gate should fire on failure trajectories but not on successful trajectories. Since the evaluation data may contain an unbalanced number of successful and failed trajectories, we measure balanced accuracy:
    \begin{equation}
    \begin{aligned}
    \mathrm{BA}(c) = \tfrac{1}{2}\big(\mathrm{TPR}(c) + \mathrm{TNR}(c)\big)
    \end{aligned}
    \label{eq:BA-formula}
    \end{equation}    
    where $c$ denotes a gate configuration. TPR corresponds to the fraction of failed trajectories where the gate fires, and TNR corresponds to the fraction of successful trajectories where the gate does not fire.
    \item  \textbf{Latency}. The gate should fire as early as possible when a failure occurs. Since it is difficult to manually label the precise onset of every failure, we use the \emph{mean trigger time} on the failed trajectories as a surrogate for detection latency. For a failure trajectory where the gate does not fire, we assign the episode horizon as its trigger time. We normalize this value by the episode horizon.
\end{itemize}
We compare \method{}'s gate against two categories of baselines: (1) \textbf{Learned gates} train additional models to produce the gating signal; this category includes ThriftyDAgger, SAFE-LSTM, and logpZO. (2) \textbf{Post-hoc gates} derive their scores directly from policies without additional offline training; Diff-DAgger and UCF fall into this category.

For each of Tasks 0, 1, 4, and 6 from LIBERO-10 suite, we collect 200 rollouts from the student policy. We reserve 50 rollouts for training the learned gates and evaluate every method on the remaining 150 rollouts. For ThriftyDAgger, we use the policy ensemble and Q-risk estimator from its final checkpoint after three rounds of DAgger, which were trained on more transitions than the 50 reserved rollouts. We train SAFE-LSTM and logpZO using the 50 rollouts according to their respective training objectives. \method{}, Diff-DAgger, and UCF require no gate training.

Comparing methods at a single operating point is insufficient because they might be noisy and difficult to compare due to the tradeoff between accuracy and trigger time. We therefore sweep multiple gate configurations and construct a Pareto frontier for each method. Let $T_{\mathrm{det}}$ be the normalized trigger time averaged across all failure trajectories. A configuration $c$ is Pareto-optimal if no other evaluated configuration $c'$ achieves both higher accuracy
$\mathrm{BA}(c')\geq\mathrm{BA}(c)$ and
lower latency $T_{\mathrm{det}}(c')\leq T_{\mathrm{det}}(c)$,
with at least one inequality strict.

To generate the candidate configurations, we sweep the score-quantile parameter $\alpha$ \cite{koenker1978regression} for each baseline. For \method{}, we sweep the false-positive-rate cap $\beta$. For each $\beta$, we choose the gate configuration from a 3600-configuration pool with the lowest trigger time subject to a false-positive rate at most $\beta$. We evaluate each configuration using five-fold cross-validation repeated over five random partitions on the 150 evaluation rollouts. Gate parameters are selected on four folds and evaluated on the held-out fold. 

\begin{figure}[t]
    \centering
    \includegraphics[
        width=\columnwidth
    ]{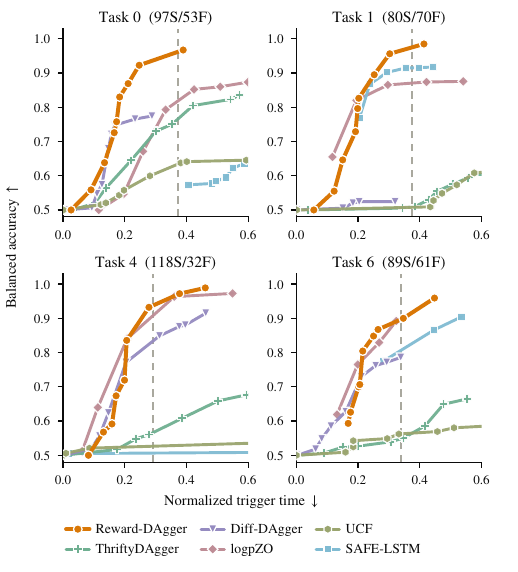}
    \vspace{-20pt}
    \caption{\textbf{Detection frontier on LIBERO tasks.} Each curve shows the
    Pareto frontier between balanced accuracy and mean normalized
    trigger time. S and F denote the numbers of successful and failed trajectories; 
    the dashed gray line indicates the mean normalized length of successful trajectories.}
    \label{fig:detection-frontier}
\vspace{-5px}
\end{figure}

The results are shown in Figure~\ref{fig:detection-frontier}. \method{} achieves \emph{the highest maximum balanced accuracy on every task}, 
reaching at least $0.96$. 
Its frontier also lies above or on par with the baselines, with large improvements on Tasks 0 and 1. logpZO remains competitive at some early-trigger operating points, but no baseline performs consistently as well as \method{} across all four tasks.

\subsection*{Q2: Transfer of Gating Configurations}
Most existing baselines use \emph{task-specific} gates, meaning that the same gating function or thresholds tuned on one task cannot be transferred directly to another. 
Several baselines are also \emph{policy-specific}, meaning that updating a policy on the same task requires retraining or recalibrating the gate. 
In contrast, \method{} depends on generic progress signals, which enable better transfer across tasks and policies. We investigate its cross-task transfer performance on LIBERO-10 tasks 0, 1, 5, and 8. 

We ask whether a single configuration tuned on one or more source tasks transfers to other tasks. 
We consider two settings: \emph{one-to-three transfer}, where we tune the gate on one task and evaluate it on the other three tasks, and \emph{three-to-one transfer}, where we tune the gate jointly on three tasks and evaluate it on the remaining task.

To select a single configuration from the source tasks, we define a balanced-accuracy tolerance $\epsilon$. Given $\epsilon$ and a set of source tasks $\mathcal{S}$, let $\mathrm{BA}_k^\star=\max_c\mathrm{BA}_k(c)$ denote the highest balanced accuracy obtained on task $k$. We select
\begin{equation}
\begin{aligned}
    c_\epsilon(\mathcal{S})
    =&\operatorname*{argmin}_{c}\;
      \frac{1}{|\mathcal{S}|}
      \sum_{k\in\mathcal{S}}T_{\mathrm{det},k}(c)\\
    &\text{subject to }\;
      \mathrm{BA}_k(c)\geq\mathrm{BA}_k^\star-\epsilon,
      \quad \forall k\in\mathcal{S}.
\end{aligned}
\label{eq:accuracy-slack}
\end{equation}
The selected configuration $c$ minimizes average trigger time while staying within $\epsilon$ of the highest balanced accuracy on every source task.

We quantify transfer using two metrics. To measure whether transfer preserves balanced accuracy, we use \emph{change in balanced accuracy}, denoted by $\Delta\mathrm{BA}=\mathrm{BA}_{\mathrm{transfer}}-\mathrm{BA}_{\mathrm{native}}$. Both quantities are measured on the target task, with $\mathrm{BA}_{\mathrm{native}}$ achieved by a configuration tuned on that task using the same $\epsilon$. To measure whether transfer preserves latency, we use \emph{matched-accuracy latency penalty}. It measures, at the same balanced accuracy, the percentage increase in trigger time relative to the configuration on the target task's native frontier. A lower penalty indicates better transfer.

\begin{table}[t]
\caption{\textbf{Configuration transfer across tolerance levels.} $\Delta\mathrm{BA}$ is in percentage points; mean, median, and worst denote matched-accuracy latency penalties (\%).}
\label{tab:single-config-transfer}
\vspace{-3px}
\centering
\footnotesize
\setlength{\tabcolsep}{3pt}
\begin{tabular}{@{}c rrrr rrrr@{}}
\toprule
& \multicolumn{4}{c}{\textbf{One-to-three}}
& \multicolumn{4}{c}{\textbf{Three-to-one}} \\
\cmidrule(lr){2-5}\cmidrule(lr){6-9}
$\epsilon$
& $\Delta\mathrm{BA}$ & \textbf{Mean} & \textbf{Median} & \textbf{Worst}
& $\Delta\mathrm{BA}$ & \textbf{Mean} & \textbf{Median} & \textbf{Worst} \\
\midrule
0.00 & $-1.6$           & 21.0 & 16.7 & 53.2 & --               & --   & --   & --   \\
0.01 & $-2.1$           & 13.1 & 12.7 & 21.9 & $-0.7$           & 21.7 & 18.9 & 33.7 \\
0.02 & $-5.0$           & 11.4 &  8.9 & 33.9 & $\phantom{-}0.0$ &  7.3 &  6.7 & 11.2 \\
0.03 & $-7.3$           & 11.6 &  8.0 & 42.9 & $\phantom{-}0.0$ &  7.6 &  7.8 & 13.7 \\
0.05 & $-7.7$           & 10.8 &  7.5 & 29.9 & $\phantom{-}0.0$ &  9.2 &  7.8 & 19.2 \\
\bottomrule
\end{tabular}
\vspace{-5px}
\end{table}

\Cref{tab:single-config-transfer} reports the results across all tested balanced-accuracy tolerance values. We note that one-to-three task transfer is not consistently reliable; although its median latency penalties are moderate, no tolerance level preserves balanced accuracy and worst-case latency. 
We attribute this to possible overfitting of task-specific sampling noise when selecting among many configurations using rollouts from only one task. In contrast, selecting the gate jointly across multiple source tasks substantially improves transfer stability, preserving average balanced accuracy over a broad range of tolerances while incurring relatively small latency penalties.

Given the reliability of three-to-one transfer, we fix $\epsilon=0.02$ and examine these configurations in greater detail. In \Cref{tab:heldout-transfer}, we report the transfer metrics of a configuration tuned on three of Tasks 0, 1, 5, and 8 and evaluated on both the omitted task and independently held-out Tasks 4 and 6. The final row reports a configuration tuned jointly on all four source tasks and evaluated on Tasks 4 and 6. 

Together, these results demonstrate that, by calibrating across a broad set of source tasks, reward gate configurations transfer reliably to target tasks, preserving balanced accuracy and inducing a small, controllable latency penalty. 

\begin{table}[t]
\caption{\textbf{Transfer results} at $\epsilon=0.02$ for three-to-one transfer and held-out Tasks 4 and 6. $T_{\mathrm{det}}$ is mean normalized trigger time; Pen is matched-accuracy latency penalty.}
\label{tab:heldout-transfer}
\vspace{-3px}
\centering
\footnotesize
\setlength{\tabcolsep}{4pt}
\begin{adjustbox}{width=\columnwidth}
\begin{tabular}{@{}lrrr rrr rrr@{}}
\toprule
& \multicolumn{3}{c}{\textbf{Three-to-one}}
& \multicolumn{3}{c}{\textbf{Eval on Task 4}}
& \multicolumn{3}{c}{\textbf{Eval on Task 6}} \\
\cmidrule(lr){2-4}
\cmidrule(lr){5-7}
\cmidrule(lr){8-10}
\textbf{Held-out task}
& $\Delta\mathrm{BA}$ & $T_{\mathrm{det}}$ & \textbf{Pen}
& $\Delta\mathrm{BA}$ & $T_{\mathrm{det}}$ & \textbf{Pen}
& $\Delta\mathrm{BA}$ & $T_{\mathrm{det}}$ & \textbf{Pen} \\
\midrule
Task 0
& $-1.2$ & 0.354 & 9.0
& $-4.4$ & 0.286 & 0.8
& $-1.7$ & 0.384 & 8.7 \\

Task 1
& $-0.9$ & 0.341 & 4.3
& $-11.0$ & 0.260 & 11.7
& $-0.8$ & 0.379 & 3.6 \\

Task 5
& $\phantom{-}1.4$ & 0.336 & 11.2
& $-1.3$ & 0.352 & 1.9
& $\phantom{-}0.0$ & 0.400 & 6.7 \\

Task 8
& $\phantom{-}0.5$ & 0.485 & 4.5
& $-1.3$ & 0.352 & 1.9
& $\phantom{-}0.0$ & 0.400 & 6.7 \\
\midrule
None (all four)
& -- & -- & --
& $-1.3$ & 0.352 & 1.9
& $\phantom{-}0.4$ & 0.412 & 3.2 \\
\bottomrule
\end{tabular}
\end{adjustbox}
\vspace{-5px}
\end{table}

\subsection*{Q3: Policy Improvement with \method{}}

\begin{figure}[t]
    \centering
    \includegraphics[
        width=\columnwidth
    ]{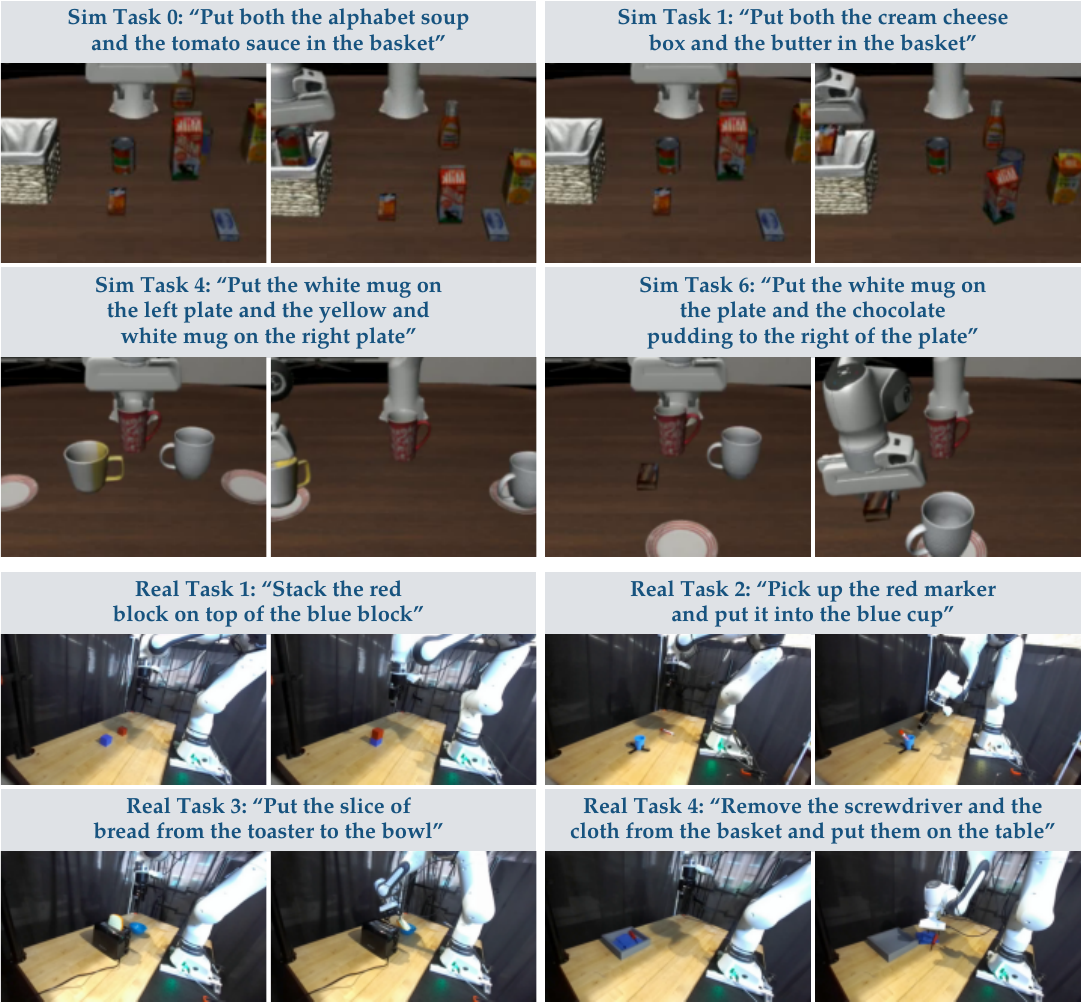}
    \caption{\textbf{Overview of simulated and real-world tasks.} Left and right images show each task’s initial and final frames.}
    \label{fig:task-overview}
\end{figure}

We next evaluate whether the interventions triggered by \method{} lead to efficient downstream policy improvement. We consider four simulated and four real-world tasks, visualized in \Cref{fig:task-overview}. Across all eight tasks, we use the same reward-gate configuration selected jointly on LIBERO Tasks 0, 1, 5, and 8:
\[
(L_s, \tau_s, \delta_{\min}, L_l, \tau_l)
= (75, -0.9, 0.05, 225, 0.0).
\]
In simulation, we query the reward model at every environment step. For real-world experiments, we address querying latency by employing asynchronous scoring and querying the reward model every $k=4$ environment steps. 

For each method, we perform policy finetuning over multiple rounds. In each round, we collect a fixed budget of expert transitions using the corresponding gating method and then finetune the base policy on a replay buffer containing the original offline demonstrations and the aggregated expert corrections. Following \citep{lee2025diff}, once takeover is triggered, the expert controls the robot for the remainder of the episode. After each round of training, we evaluate two metrics:
\begin{itemize}[leftmargin=*]
     \item \textbf{Autonomous success rate} evaluates the updated student policy over $K$ autonomous rollouts. It reflects how effectively the gate triggers informative expert interventions.
     \item \textbf{Return on human effort (ROHE)} evaluates the performance of the combined student-expert system relative to the amount of expert interventions required. Following \cite{liu2024modelbased}, ROHE is defined as
    \begin{equation}
        \mathrm{ROHE} = \frac{\mathbb{P}(S_\tau=1)}{1 + H/T}.
    \end{equation}
    The numerator corresponds to the task's success rate when the gate is active and the expert can take over. $H/T$ denotes the fraction of rollout episodes in which the expert intervenes. A high ROHE indicates that the combined system achieves a high success rate while requiring expert intervention in relatively few episodes. We measure ROHE at the episode level as each intervention in our setting results in expert takeover for the remainder of the episode.
\end{itemize}

\begin{figure*}[t]
    \centering
    \includegraphics[
        width=0.85\textwidth
     ]{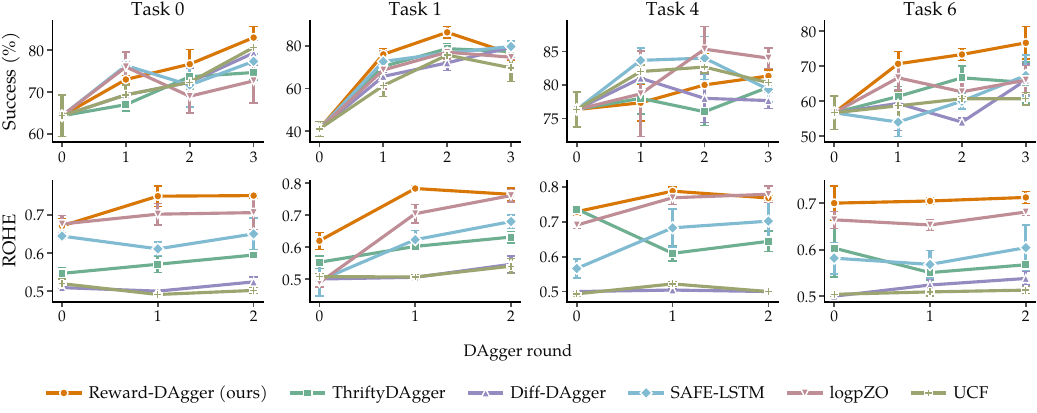}
     \vspace{-5pt}
    \caption{
  \textbf{Policy improvement results} on LIBERO-10 Tasks 0, 1, 4, and 6. 
  Error bars indicate standard error across 3 seeds. 
  }
    \label{fig:sim-success-rohe}
    \vspace{-3px}
\end{figure*}

\noindent\textbf{Simulation Results.} We evaluate \method{} on LIBERO-10 tasks 0, 1, 4, and 6. 
Among the downstream tasks, Tasks 0 and 1 were included in gate-configuration tuning, whereas Tasks 4 and 6 were held out. 

We perform three rounds of interactive learning and collect 3,000 successful expert state-action pairs for each round. We compute each round's ROHE from the expert-gated rollouts during data collection. We then finetune the current student policy using a sampling mixture of $30\%$ offline demonstrations and $70\%$ aggregated expert interventions. After each round, we evaluate the updated policy's autonomous success rate using 100 rollouts. Since the final policy is not used in a subsequent collection round, we do not report ROHE for the final checkpoint. 

Because UCF, SAFE, and logpZO were not originally designed for iterative policy updates, we collect 50 additional rollouts from the updated student policy after each round. For UCF, these rollouts are used to recalibrate its uncertainty threshold; for SAFE and logpZO, they are used to retrain their scoring models and recalibrate their thresholds. 

The results, averaged across three seeds, are reported in \Cref{fig:sim-success-rohe}. \method{} improves autonomous success on all four tasks, achieving the highest final success on Tasks 0 and 6 and remaining competitive on Tasks 1 and 4.
It also achieves the highest or near-highest ROHE throughout data collection, indicating that these improvements are obtained with efficient use of expert interventions. Importantly, these gains extend to Tasks 4 and 6 despite neither task being used to select the gate configuration.

\begin{table}[t]
\caption{\textbf{Real-world results.} 
}
\label{tab:real-results}
\vspace{-3px}
\centering
\small
\setlength{\tabcolsep}{4pt}
\begin{tabular}{@{}ll ccc ccc@{}}
\toprule
& & \multicolumn{3}{c}{Success rate (\%)} & \multicolumn{3}{c}{ROHE} \\
\cmidrule(lr){3-5} \cmidrule(l){6-8}
Task & Method & R0 & R1 & R2 & R0 & R1 & R2 \\
\midrule
Stack & \rewardcolor{\method{}} & 65 & \textbf{75} & \textbf{80} & \textbf{.667} & .731 & \textbf{.643} \\
 & \thriftycolor{ThriftyDAgger} & 65 & 70 & 55 & .606 & .556 & .513 \\
 & \diffcolor{Diff-DAgger} & 65 & \textbf{75} & 70 & .588 & \textbf{.783} & .600 \\
\midrule
Pick & \rewardcolor{\method{}} & 20 & 30 & \textbf{40} & .541 & \textbf{.533} & \textbf{.588} \\
 & \thriftycolor{ThriftyDAgger} & 20 & 25 & 25 & .526 & .474 & .424 \\
 & \diffcolor{Diff-DAgger} & 20 & \textbf{35} & \textbf{40} & \textbf{.548} & .435 & .478 \\
\midrule
Put & \rewardcolor{\method{}} & 35 & \textbf{40} & 45 & \textbf{.594} & \textbf{.606} & \textbf{.645} \\
 & \thriftycolor{ThriftyDAgger} & 35 & 35 & \textbf{50} & .559 & .500 & .526 \\
 & \diffcolor{Diff-DAgger} & 35 & 20 & 30 & .500 & .346 & .528 \\
\midrule
Remove & \rewardcolor{\method{}} & 75 & \textbf{85} & \textbf{90} & \textbf{.588} & \textbf{.714} & .769 \\
 & \thriftycolor{ThriftyDAgger} & 75 & 70 & 80 & .526 & .486 & .563 \\
 & \diffcolor{Diff-DAgger} & 75 & 70 & 80 & .545 & .613 & \textbf{.773} \\
\bottomrule
\end{tabular}
\vspace{-5px}
\end{table}

\noindent\textbf{Real-robot Tasks.} We then evaluate \method{} on four real-robot tasks.
For \emph{stack}, \emph{pick}, and \emph{put}, there exist exact or closely related examples in the dataset used to train Robometer-4B. In contrast, the \emph{remove} task involves manipulating two distinct objects in sequence and serves as an out-of-distribution (OOD) task.

We use $\pi_{0.5}$-DROID as the student policy and provide real human interventions through teleoperation. We deploy the same reward gate configuration selected in simulation without additional tuning on the real-world tasks. We compare \method{} with ThriftyDAgger and Diff-DAgger and evaluate every iteration's policy checkpoint with 20 rollouts. We exclude UCF, SAFE, and logpZO from the real-world comparisons because these methods require obtaining fresh rollouts of the latest policy for recalibration or retraining. 
Under our 20-rollout evaluation budget, we find that calibration is highly unstable. 
To compute ROHE, we use the evaluation rollouts, where the human annotator judges whether the state at which the gate triggers is recoverable. A rollout is considered successful for the combined student-expert system if the student completes the task autonomously or the gate triggers while expert recovery remains feasible. 

Results are reported in \Cref{tab:real-results}. 
\method{} achieves strong overall performance, obtaining the highest ROHE in most settings and comparable performance in the others, while consistently improving autonomous success across DAgger rounds.
This advantage extends to the \emph{remove} task, demonstrating Reward DAgger's effectiveness on OOD tasks. Empirically, ThriftyDAgger and Diff-DAgger frequently trigger during successful rollouts, particularly in early rounds, leading to unnecessary interventions; in contrast, \method{} rarely triggers on successful behavior. 

Overall, our results show that \emph{\method{} collects interventions that are both informative for downstream policy improvement and efficient in their use of human supervision}. Notably, the same reward model and gate configuration is used without additional tuning across eight simulated and real-world tasks, and different policy architectures.

\subsection*{Ablations}

We conduct two ablations to evaluate our short- and long-window gate design and whether \method{} remains effective across different reward models.

\begin{table}[t]
\caption{\textbf{Gate design ablation:} earliest normalized trigger time for each balanced-accuracy threshold.}
\label{tab:gate-ablation}
\vspace{-3px}
\centering\small\setlength{\tabcolsep}{5pt}
\begin{tabular}{@{}ll cccc@{}}
\toprule
& & \multicolumn{4}{c}{Trigger time at balanced accuracy $\geq$} \\
\cmidrule(l){3-6}
Task & Detector & 0.80 & 0.85 & 0.90 & 0.95 \\
\midrule
\multirow{4}{*}{Task 0} & Full gate & \textbf{.184} & \textbf{.211} & \textbf{.224} & \textbf{.389} \\
 & Short only & .194 & .228 & .255 & -- \\
 & Long only & .219 & .219 & .234 & .411 \\
 & Absolute thresh. & .205 & .266 & .331 & .500 \\
\midrule
\multirow{4}{*}{Task 1} & Full gate & \textbf{.199} & .252 & \textbf{.261} & \textbf{.303} \\
 & Short only & .233 & \textbf{.241} & .299 & .324 \\
 & Long only & .243 & .260 & .279 & .319 \\
 & Absolute thresh. & .270 & .270 & .323 & .385 \\
\midrule
\multirow{4}{*}{Task 4} & Full gate & \textbf{.206} & \textbf{.212} & \textbf{.279} & .378 \\
 & Short only & .342 & .342 & -- & -- \\
 & Long only & .242 & .312 & .337 & .373 \\
 & Absolute thresh.& .313 & .313 & .313 & \textbf{.357} \\
\midrule
\multirow{4}{*}{Task 6} & Full gate & \textbf{.215} & .265 & .347 & .448 \\
 & Short only & \textbf{.215} & \textbf{.249} & .358 & -- \\
 & Long only & .319 & .319 & \textbf{.329} & .551 \\
 & Absolute thresh. & .287 & .314 & .355 & \textbf{.445} \\
\bottomrule
\end{tabular}
\vspace{-5px}
\end{table}

\noindent\textbf{Gate Design.} We compare the complete reward gate against three baselines: a short-window-only gate, a long-window-only gate, and an absolute-progress gate, which triggers when progress drops below a certain amount. 

Table~\ref{tab:gate-ablation} summarizes the resulting accuracy-latency tradeoff. The full gate achieves the lowest or tied trigger time in 11 of the 16 task-accuracy combinations, highlighting the importance of combining both gates. The short-window-only gate detects some failures earlier but fails to reach high balanced accuracy on Tasks 0, 4, and 6, whereas the long-window-only gate provides broader coverage at the cost of later detection. The full gate also triggers earlier than the absolute-progress gate in 14 of the 16 comparisons.

\begin{figure}[t]
    \centering
    \includegraphics[
        width=\columnwidth
    ]{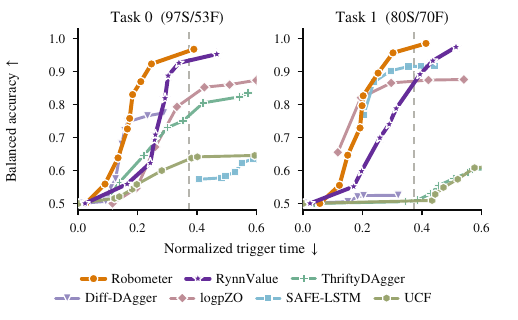}
    \vspace{-20pt}
    \caption{\textbf{Reward model ablation.} Detection frontiers using Robometer and RynnValue as reward model backbones.}
    \label{fig:ablation}
    \vspace{-5px}
\end{figure}

\noindent\textbf{Different Reward Model.}
To test whether our gating mechanism generalizes across progress-based reward models, we replace Robometer-4B with RynnValue-4B while keeping the gate construction and evaluation protocol unchanged. Fig.~\ref{fig:ablation} shows that, under the same configuration-selection procedure on two simulation tasks, RynnValue achieves a comparable accuracy--latency frontier and the second-highest balanced accuracy on both tasks, suggesting that the gating mechanism is not specific to a particular reward-model backbone.

\section{Conclusion}

\noindent\textbf{Summary.} We introduced Reward-DAgger, a robot-gated interactive imitation learning framework that detects progress drops and plateaus using estimates from a general-purpose reward model. By decoupling the gating mechanism from the underlying policy, Reward-DAgger operates across policy checkpoints and architectures, including black-box policies, without retraining or recalibrating the gate between DAgger rounds. In simulated and real-robot experiments, \method{} achieves effective failure detection, cross-task transfer, and efficient downstream policy improvement. \method{} is modular with respect to the reward model: more accurate general-purpose reward models can be incorporated without modifying the interactive imitation learning pipeline. 

\noindent\textbf{Limitations and Future Work.} 
\method{} is not without limitations. Its long-window gate must observe a sustained progress plateau, so failures such as repeated unsuccessful grasp attempts may persist for several seconds before triggering an intervention. More broadly, the framework is reactive rather than preemptive, requesting human assistance only after evidence of a failure emerges, which limits its applicability in high-risk settings or when failures are difficult to recover from. A promising direction is to integrate predictive world models \citep{yin2026playworld, wang2026interactive} into the gating framework, allowing the robot to anticipate possible failure outcomes and request intervention before executing them.

\section*{Acknowledgments}
The authors thank the USC Center for Advanced Research Computing (CARC) for providing us with compute resources, and acknowledge funding by the Airbus Institute for Engineering Research (AIER). This work was supported by the Office of Naval Research (ONR) Young Investigator Program (YIP) under grant number N00014-26-1-2473.

\balance
\printbibliography
\newpage
\onecolumn
\section*{Supplementary Material}

\subsection*{Additional Experiments}

\noindent\textbf{Frontier Transfer.} We explore whether configurations that perform well on one task remain effective on other tasks. For each source task, we construct its task-specific Pareto frontier using 200 policy rollouts from a diffusion policy trained exclusively on that task. We then apply the configurations on this frontier to each of the other three tasks and retain the Pareto-optimal points to form a transferred frontier.

To measure how closely each transferred frontier matches the target task's native frontier, we compute the hypervolume ratio (HVR). Specifically, we divide the area under the transferred frontier by the area under the native frontier. The trigger-time range extends from zero to the latest non-dominated point reached by either frontier. An HVR near one indicates that the transferred configurations retain most of the accuracy--latency tradeoff obtained by task-specific tuning.

\begin{figure}[h]
    \centering
    \includegraphics[
        width=0.5\columnwidth
    ]{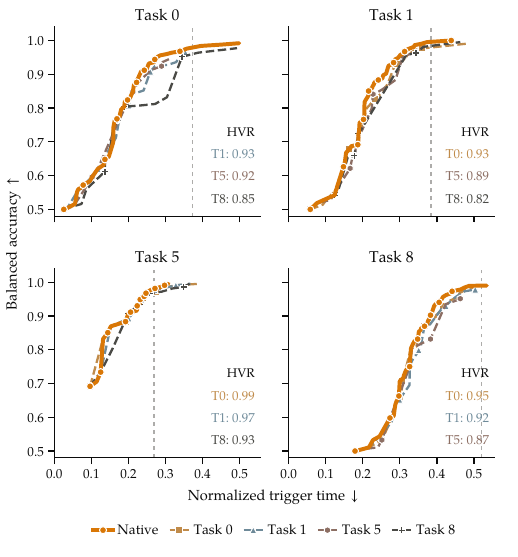}
    \caption{\textbf{Frontier transfer of \method{} gate configurations.} Each panel corresponds to a target task. The solid orange curve denotes the native frontier obtained through task-specific tuning, and each dashed curve is obtained by applying configurations from another task's frontier without retuning. Numbers indicate the hypervolume ratios (HVR) for each transferred curve. The dashed grey line indicates the mean normalized length of successful trajectories.}
    \label{fig:gate-frontier-hv}
\end{figure}

The results for each source-target task pair are shown in \Cref{fig:gate-frontier-hv}. Across the 12 ordered source-target pairs, \method{} achieves a mean HVR of $0.92$ and a median HVR of $0.93$, indicating that \emph{high-performing regions of the gate configuration space are largely shared across tasks.}

\noindent\textbf{Different Reward Model Backbone in Real World Tasks.} We further replace Robometer with RynnValue on the \emph{Stack} and \emph{Pick} policy-improvement experiments while directly reusing the Robometer-selected gating parameters, without additional tuning. As shown in \Cref{tab:rynnvalue-ablation}, RynnValue yields smaller gains in autonomous success but comparable ROHE overall, indicating that the proposed gating mechanism is not specific to a particular reward-model backbone.

\begin{table}[h]
\caption{\textbf{Different reward model backbone ablation on real-world tasks.}}
\label{tab:rynnvalue-ablation}
\vspace{-3px}
\centering
\small
\setlength{\tabcolsep}{4pt}
\renewcommand{\cellalign}{cl}
\begin{tabular}{@{}ll ccc ccc@{}}
\toprule
& & \multicolumn{3}{c}{Success rate (\%)} & \multicolumn{3}{c}{ROHE} \\
\cmidrule(lr){3-5} \cmidrule(l){6-8}
Task & Method & R0 & R1 & R2 & R0 & R1 & R2 \\
\midrule
Stack & Robometer-4B & 65 & \textbf{75} & \textbf{80} & \textbf{.667} & \textbf{.731} & \textbf{.643} \\
 & RynnValue-4B & 65 & 65 & 70 & .606 & .645 & .613 \\
\midrule
Pick & Robometer-4B & 20 & \textbf{30} & \textbf{40} & \textbf{.541} & .533 & \textbf{.588} \\
 & RynnValue-4B & 20 & 25 & 25 & .515 & \textbf{.556} & .571 \\
\bottomrule
\end{tabular}
\vspace{-5px}
\end{table}

\subsection*{Implementation Details}

\begin{table*}[h]
\caption{Hyperparameter grids for detection frontier analysis}
\label{tab:gate-sweeps}
\centering
\small
\setlength{\tabcolsep}{5pt}

\begin{tabular}{@{}lllp{6.2cm}c@{}}
\toprule
\textbf{Method} &
\textbf{Role} &
\textbf{Parameter} &
\textbf{Values} &
\textbf{Grid size} \\
\midrule

\multirow{6}{*}{Reward-DAgger}
& Frontier
& FPR cap $\beta$
& $\mathcal{B}$
& 22 \\

\cmidrule(lr){2-5}

& \multirow{5}{*}{Inner search}
& Short-window length $L_s$
& $\{20,\,30,\,50,\,75,\,100\}$
& \multirow{5}{*}{3,600} \\

&
& Short-window threshold $\tau_s$
& $\{-0.9,\,-0.7,\,-0.5\}$
& \\

&
& Minimum drop $\delta_{\min}$
& $\{0,\,0.05,\,0.10,\,0.20\}$
& \\

&
& Long-window length $L_l$
& $\{150,\,175,\,\ldots,\,400,\,450\}$
& \\

&
& Long-window threshold $\tau_l$
& $\{-0.2,\,0,\,0.1,\,0.3,\,0.5\}$
& \\

\midrule

ThriftyDAgger
& Frontier
& Novelty/Q-risk quantile $\alpha$
& $\mathcal{A}$
& 43 \\

Diff-DAgger
& Frontier
& Diffusion-loss quantile $\alpha$
& $\mathcal{A}$
& 43 \\

UCF
& Frontier
& Uncertainty quantile $\alpha$
& $\mathcal{A}$
& 43 \\

SAFE-LSTM
& Frontier
& Score quantile $\alpha$
& $\mathcal{A}$
& 43 \\

logpZO
& Frontier
& Score quantile $\alpha$
& $\mathcal{A}$
& 43 \\

\midrule

\multicolumn{5}{@{}l}{
\footnotesize
$\mathcal{B}
=\{0,\,0.02\}\cup\{0.05,\,0.10,\,\ldots,\,1.00\}$.} \\

\multicolumn{5}{@{}l}{
\footnotesize
$\mathcal{A}
=\operatorname{geomspace}(10^{-4},0.5,41)
\cup\{10^{-3},10^{-2}\}$.} \\

\bottomrule
\end{tabular}
\end{table*}

\begin{table*}[h]
\caption{Training and evaluation details}
\label{tab:setup}
\centering
\small
\renewcommand{\arraystretch}{1.15}
\setlength{\tabcolsep}{3pt}

\begin{tabularx}{\textwidth}{
@{}
>{\centering\arraybackslash}p{0.23\textwidth}
*{8}{Y}
@{}}
\toprule
& \multicolumn{4}{c}{\textbf{Simulated tasks}}
& \multicolumn{4}{c}{\textbf{Real-robot tasks}} \\
\cmidrule(lr){2-5}\cmidrule(l){6-9}

& Task 0 & Task 1 & Task 4 & Task 6
& Stack & Pick & Put & Remove \\
\midrule

Successful offline transitions
& 9,293 & 12,702 & 10,353 & 9,245
& 4,265 & 3,536 & 3,436 & 7,454 \\

\midrule
Observations
& \multicolumn{4}{c}{
  \makecell{Agent-view RGB, wrist RGB,\\and proprioception}}
& \multicolumn{4}{c}{
  \makecell{Exterior RGB, wrist RGB, joint positions (7),\\
  and gripper state (1)}} \\[2pt]

Image crop size
& \multicolumn{8}{c}{$224 \times 224$ (resize with padding)} \\

\midrule
Action chunk (predicted/executed)
& \multicolumn{4}{c}{\raisebox{-0.7\height}{20/10}}
& \multicolumn{4}{c}{\raisebox{-0.7\height}{15/8}} \\

Control frequency
& \multicolumn{4}{c}{20\,Hz}
& \multicolumn{4}{c}{15\,Hz} \\

Episode horizon
& \multicolumn{4}{c}{800 steps}
& \multicolumn{4}{c}{900 steps} \\

\midrule
Student policy
& \multicolumn{4}{c}{
  \makecell{Diffusion policy trained on \\ offline expert demonstrations}}
& \multicolumn{4}{c}{$\pi_{0.5}$-DROID} \\[2pt]

Data Aggregation
& \multicolumn{4}{c}{\makecell{3 rounds, 3000 transitions per round}}
& \multicolumn{4}{c}{\makecell{2 rounds, 1500 transitions per round}} \\

Training per round
& \multicolumn{4}{c}{\makecell{Full FT, 4k steps, batch size 256, \\ constant lr 1e-4}}
& \multicolumn{4}{c}{\makecell{LoRA, 3k steps, batch size 32, \\ cosine lr $2.5{\times}10^{-5}$}} \\

Expert
& \multicolumn{4}{c}{$\pi_{0.5}$-LIBERO}
& \multicolumn{4}{c}{Human teleoperator} \\

Evaluation rollouts per round
& \multicolumn{4}{c}{100}
& \multicolumn{4}{c}{20} \\
\bottomrule
\end{tabularx}
\end{table*}

\end{document}